\documentclass[conference]{IEEEtran}
\IEEEoverridecommandlockouts
\usepackage{cite}
\usepackage{amsmath,amssymb,amsfonts}
\usepackage{algorithmic}
\usepackage{graphicx}
\usepackage{multirow}
\usepackage[ruled,vlined]{algorithm2e}
\usepackage{color}
\usepackage{textcomp}
\usepackage{xcolor}
\usepackage{booktabs}
\usepackage[colorlinks,bookmarksopen,bookmarksnumbered,citecolor=green, linkcolor=red, urlcolor=green]{hyperref}
\def\BibTeX{{\rm B\kern-.05em{\sc i\kern-.025em b}\kern-.08em
    T\kern-.1667em\lower.7ex\hbox{E}\kern-.125emX}}
\begin{document}

\title{Pruning then Reweighting: Towards Data-Efficient Training of Diffusion Models \thanks{\footnotesize \textsuperscript{*}Note: Under Review}}
% *\\
% {\footnotesize \textsuperscript{*}Note: Sub-titles are not captured for https://ieeexplore.ieee.org  and
% should not be used}
% \thanks{Identify applicable funding agency here. If none, delete this.}

\author{Yize Li$^1$, Yihua Zhang$^2$, Sijia Liu$^{2}$, Xue Lin$^1$\\$^1$ Department of Electrical and Computer Engineering, Northeastern University, Boston, USA\\\{li.yize, xue.lin\}@northeastern.edu\\$^2$ Department of Computer Science and Engineering, Michigan State University, East Lansing, USA\\\{zhan1908, liusiji5\}@msu.edu}

\maketitle

\begin{abstract}
Despite the remarkable generation capabilities of Diffusion Models (DMs), conducting training and inference remains computationally expensive. Previous works have been devoted to accelerating diffusion sampling, but achieving data-efficient diffusion training has often been overlooked. In this work, we investigate efficient diffusion training from the perspective of dataset pruning. Inspired by the principles of data-efficient training for generative models such as generative adversarial networks (GANs), we first extend the data selection scheme used in GANs to DM training, where data features are encoded by a surrogate model, and a score criterion is then applied to select the coreset. To further improve the generation performance, we employ a class-wise reweighting approach, which derives class weights through distributionally robust optimization (DRO) over a pre-trained reference DM. For a pixel-wise DM (DDPM) on CIFAR-10, experiments demonstrate the superiority of our methodology over existing approaches and its effectiveness in image synthesis comparable to that of the original full-data model while achieving the speed-up between 2.34$\times$ and 8.32$\times$. Additionally, our method could be generalized to latent DMs (LDMs), e.g., Masked Diffusion Transformer (MDT) and Stable Diffusion (SD), and achieves competitive generation capability on ImageNet. Code is available \href{https://github.com/Yeez-lee/Data-Selection-and-Reweighting-for-Diffusion-Models}{here}.
\end{abstract}


\section{Introduction}
Diffusion Models (DMs)~\cite{ho2020denoising,song2021scorebased,dhariwal2021diffusion, rombach2021highresolution} belong to a recent class of generative models,  which have achieved state-of-the-art generation performance~\cite{Peebles2022DiT, ho2022video, li2023layerdiffusion, xin2024parameter, zhang2023adding, liang2023flowvid, li2024ecnet, li2024tuning}. Despite their superiority in terms of training stability, versatility, and scalability, DMs are known for their slow generation speeds due to the requirement of reverse diffusion processing by passing through the generator at massive times. Consequently, there is considerable interest in enhancing the inference speed of DMs~\cite{salimans2022progressive, meng2023distillation, li2024snapfusion, liu2023instaflow}. Furthermore, DMs are recognized for their high training costs. Modeling complicated and high-dimensional data distributions requires numerous iterations, resulting in exponential growth in training costs under the increasing resolution and diversity of the data. 

Several works have considered speeding up diffusion training by the progressive patch size~\cite{wang2023patch}, masked patches~\cite{Zheng2024MaskDiT, ding2024patched}, momentum stochastic gradient descent (SGD)~\cite{wu2023fast} and a clamped signal-to-noise ratio (SNR) weight at time-step~\cite{Hang_2023_ICCV}.
However, none of them attempted to achieve efficient training through the lens of dataset pruning (or coreset selection). To the best of our knowledge, this is the first work to investigate how the coreset size of training data influences the generation ability of DMs. In this study, we first utilize a GAN-based data selection method~\cite{devries2020instance} for diffusion training, which consists of feature embedding and data scoring. To refine training data distribution, a perceptually aligned embedding function~\cite{zhang2018unreasonable}, such as the latent space of a pre-trained image classiﬁer (e.g., Inceptionv3~\cite{szegedy2015rethinking}) is to acquire the data feature space. Then, a scoring criterion (e.g., Gaussian model) is to rank each data point in the embedding space and remove less relevant data. Nevertheless, we discover that such a data selection approach may generalize poorly to DMs on small-scale datasets.
Hence, there is a pressing need for innovations to enhance the current data selection scheme for diffusion-based generative models.

\begin{figure}[htbp]
  \centering
  % \fbox{\rule{0pt}{2in} \rule{1.0\linewidth}{0pt}}
   \includegraphics[width=1.0\linewidth]{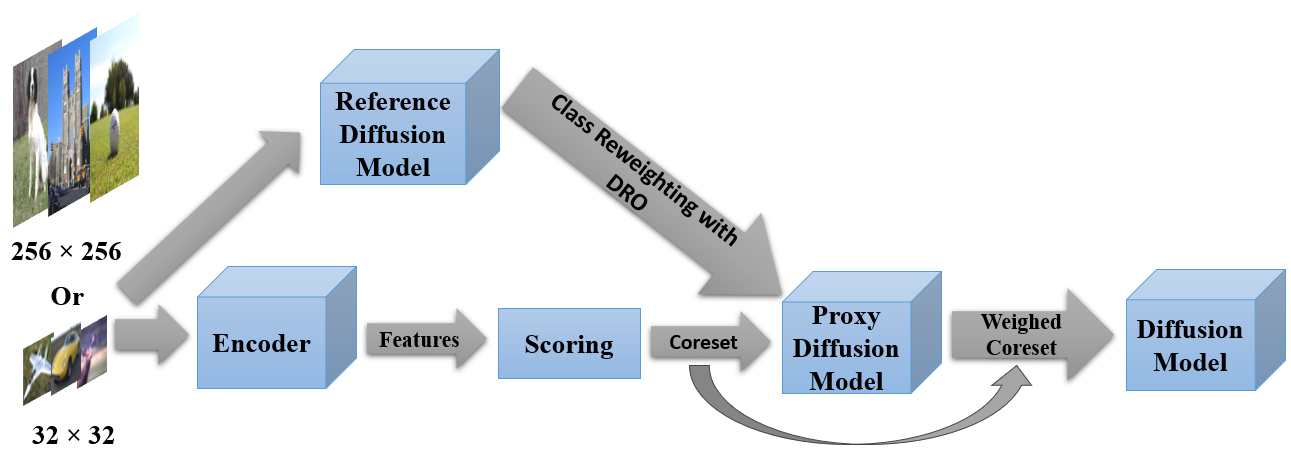}
   \caption{Overview of our Data-Efficient Diffusion Training approach: Given input images (e.g., image size of 32$\times$32 for DDPM~\cite{ho2020denoising} and 256$\times$256 for MDT~\cite{gao2023masked} and SD~\cite{rombach2021highresolution}), we use a surrogate model as an encoder to obtain latent features and the following scoring function is to prune the datasets. The pre-trained reference DM facilitates the training of a proxy DM using distributionally robust optimization (DRO ~\cite{sagawa2020distributionally, xie2023doremi}) across classes to generate class weights. Subsequently, DMs are trained on the weighted subset. }
   \label{fig:fig1}
\end{figure}

We summarize our proposed pipeline in Fig.~\ref{fig:fig1}, which investigates the encoder and scoring method to implement data selection in DMs. 
Inceptionv3~\cite{szegedy2015rethinking}, ResNet-18~\cite{He_2016_CVPR}, CLIP~\cite{radford2021learning} and DDAE~\cite{ddae2023} are adopted as the choices of surrogate models (encoder) and the scoring functions (dataset pruning methods) are Gaussian model~\cite{ devries2020instance} and Moderate-DS~\cite{xia2023moderate}, which keep data points with scores within the scoring threshold. One key observation is that simply pruning the dataset might lower generation capability, with the generative capacity of each class decreasing to varying extents in Fig.~\ref{fig:fig2}. To address this issue, we leverage a class-wise reweighting strategy by distributionally robust optimization (DRO~\cite{sagawa2020distributionally, xie2023doremi}), to optimize the class weights that are dynamically updated according to the marginal loss on each class. Experimental results on the pixel-level DDPM~\cite{ho2020denoising}, the latent-level Masked Diffusion Transformer (MDT)~\cite{gao2023masked} and Stable Diffusion (SD)~\cite{rombach2021highresolution} demonstrate that our method could accelerate diffusion training from \textbf{2.34$\times$} up to \textbf{8.32$\times$} while maintaining comparable or even superior generation ability. The main contributions are highlighted below.
\begin{itemize}
    \item We investigate the problem of efficient DM training through the lens of  dataset pruning for the first time, which selects coreset from the latent space through surrogate models.
    \item We develop a novel class-wise reweighting strategy to enhance generation capacity by minimizing the variance between the target proxy model and the reference model.
    \item We achieve comparable performances on DDPM and notable sampling improvements on latent diffusion models (LDMs) while obtaining gains in computation efficiency.
\end{itemize}

\section{Related Work}
\label{sec:relatedwork}

\begin{table*}[t]
\caption{CIFAR-10 FID results (32$\times$32) for classifier-free guided~\cite{ho2022classifierfree} DDPM~\cite{ho2020denoising} by DDIM sampler~\cite{song2020denoising} on [10\%, 20\%, 30\%, 40\%, 100\%] of training data. With ResNet-18 as the encoder and class-wise reweighting, our approach significantly outperforms other baselines and is comparable to that under full dataset training. }
  \centering
  \begin{tabular}{c|c|ccccc}
    \toprule[1pt]
     \multirow{2}*{Surrogate Model} & \multirow{2}*{Selection Method} & \multicolumn{5}{c}{FID ($\downarrow$) under Data Ratio}\\
     & & 10\% & 20\% & 30\% & 40\% & 100\%\\
    \midrule
    N/A & Uniform Random & 8.12 & 5.64 &4.86 &4.45 &\multirow{6}*{3.66}\\
    Inceptionv3 & Gaussian & 22.03 &15.65 &11.70  & 9.34  & \\
    DDAE & Moderate-DS & 7.54  &5.35 &4.69 &4.58 & \\
    CLIP & Moderate-DS & 7.78  &5.89 &5.04 &4.93 &\\
    \multirow{2}*{ResNet-18} & Moderate-DS & 7.39 &5.26 &4.54 &4.31 &\\
    & + Reweighting & \textbf{6.71}  &\textbf{4.95} &\textbf{4.44} &\textbf{4.18} &\\
    \bottomrule
  \end{tabular}
  \label{tab:tab1}
\end{table*}

Diffusion models~\cite{ho2020denoising,song2021scorebased,dhariwal2021diffusion, esser2023structure} are proposed to capture the high-dimensional nature of data distributions, which are dominating a new era by exceeding the Generative adversarial networks (GANs)~\cite{karras2019stylebased, Sauer2021ARXIV}. The backbone networks of DMs generally include the convolutional U-Net~\cite{10.1007/978-3-319-24574-4_28}, 
% which consists of a contracting path and an expanding path that enables precise localization in a symmetrical form, 
and the transformer-based architectures~\cite{Peebles2022DiT, dosovitskiy2020image, bao2022all} with attention layers.

\subsection{Efficiency in Diffusion Models}
The sampling of DMs is typically costly because of the iterative denoising process with UNet and the DM training is always time-consuming by massive steps. To address these issues, existing works concentrate on reducing sampling steps through step distillation~\cite{salimans2022progressive, meng2023distillation, yin2024onestep} and efficient sampling solvers, including DDIM~\cite{song2020denoising} and DPM-Solver~\cite{lu2022dpm}. Other recent works consider compression~\cite{li2024snapfusion, fang2024structural} and utilize the property of the model architecture~\cite{ma2023deepcache, si2023freeu}. Furthermore, accelerating diffusion training is achieved by gradually scaling up image size~\cite{wang2023patch}, or token merging and masking~\cite{gao2023masked, wei2023diffusion, Zheng2024MaskDiT, ding2024patched} in transformer-based DMs.

\subsection{Dataset Pruning}
Dataset pruning, also known as coreset selection, refers to reducing training data by creating a more compact dataset~\cite{borsos2020coresets, li2023more}. 
% Proxy functions~\cite{Kaushal_2019_WACV, coleman2020selection} leverage the feature representation from a smaller proxy model to identify the most informative subset for training the larger model. 
A small representative subset can be approximated based on training dynamics as the score criterion~\cite{paul2021deep, yang2023dataset}, and loss or gradient perspectives, such as GRAD-MATCH~\cite{pmlr-v139-killamsetty21a}, RHO-LOSS~\cite{mindermann2022prioritized} and InfoBatch~\cite{qin2024infobatch}.

\section{Methodology}
\label{sec:method}
\subsection{Background}
\noindent \textbf{Diffusion models.} DMs include a forward noising process and a backward denoising process to estimate the distribution of data iteratively~\cite{ho2020denoising, song2020denoising, song2021scorebased, rombach2021highresolution}. Given the clean input $\boldsymbol{x}_0$, it is gradually turned into the noisy $\boldsymbol{x}_T$ over $T$ time steps ($\boldsymbol{x}_t$ at each time step $t$) by Gaussian noise $\boldsymbol{\epsilon}\sim \mathcal{N}(\mathbf{0}, \mathbf{I})$ in the forward diffusion process. In the backward sampling process, a noisy sample $\boldsymbol{x}_T \sim \mathcal{N}(\mathbf{0}, \mathbf{I})$ is progressively denoised to generate an uncorrupted output. The objective of DM can be simplified by minimizing the noise approximation error
\begin{equation}
\mathbb{E}_{\boldsymbol{x}, c, \boldsymbol{\epsilon}, t}[\|\boldsymbol{\epsilon}_{\boldsymbol{\theta}}(\boldsymbol{x}_{t}, c, t)-\boldsymbol{\epsilon}\|^{2}],
\label{eq:eq1}   
\end{equation}%
where $\boldsymbol{\epsilon}_{\boldsymbol{\theta}}(\boldsymbol{x}_{t}, c, t)$ represents the noise estimator at time step $t$ over trainable parameter $\boldsymbol \theta$ regarding with the condition $c$ (e.g., class label or text prompt). In conditional DDPM~\cite{ho2020denoising}, $\boldsymbol{x}_t$
denotes the input image, while in latent diffusion model (LDM)~\cite{rombach2021highresolution}, $\boldsymbol{x}_t$
is the latent feature. 
% \noindent \textbf{Classifier-free guidance.} 
Classifier-free guidance~\cite{ho2022classifierfree} has been demonstrated to significantly enhance the
sample quality of class-conditioned DMs. Specifically, a guidance weight $w \geq 0$ is introduced to balance generation quality and sample diversity, where a conditional DM with the condition $c$ is jointly trained with an unconditional DM. The new noise estimation from Eq.~\eqref{eq:eq1} is formulated as $\boldsymbol{\epsilon}^{w}_{\boldsymbol{\theta}}(\boldsymbol{x}_{t}, c, t) = (1+w)\boldsymbol{\epsilon}_{\boldsymbol{\theta}}(\boldsymbol{x}_{t}, c, t)-w\boldsymbol{\epsilon}_{\boldsymbol{\theta}}(\boldsymbol{x}_{t}, t)$.

\subsection{Dataset Pruning}
% \noindent \textbf{Representation encoding.} 
Dataset pruning consists of two parts, embedding by a surrogate model and ranking by a scoring function.
Given a pre-trained network $f(\boldsymbol{x}, y)=g(h(\boldsymbol{x}, y))$, where
$h$ means the encoder that converts inputs to latent representations, and $g$ is the classification head. $\boldsymbol{x}_i\in \mathbb{R}^{d}$ and $y_i\in[K]$ are the input sample and its corresponding label, respectively. The representations $\boldsymbol{z}$ are obtained as $\{\boldsymbol{z}_{1}=h({\boldsymbol{x}}_{1}, {y}_{1}), \ldots,\boldsymbol{z}_{n}=h(\boldsymbol{x}_{n}, y_{n})\}$. The Gaussian model~\cite{devries2020instance} then computes a score for each embedded sample $\boldsymbol{z}_{i}$ by the empirical mean $\boldsymbol{\mu}$ and the covariance $\boldsymbol{\Sigma}$, which is defined by $S_{\mathrm{Gaussian}}(\boldsymbol{z}_i) = -\frac{1}{2}\left [(\boldsymbol{z}_i - \boldsymbol{\mu})^{T}\boldsymbol\Sigma^{-1}(\boldsymbol{z}_i - \boldsymbol{\mu})  \right. \nonumber + \ln(|\boldsymbol\Sigma|) 
 + \left.  d \ln(2\pi) \right. ]$,
where $d$ denotes the dimension of $\boldsymbol{z}_i$. After scoring each sample, we preserve all data points with scores larger than a threshold $\tau$, which equals a certain percentile of the scores, ensuring that the top $N$\% (data ratio $R$) of the highest-scoring data points are retained to formulate a coreset $D_s$.

Moderate-DS~\cite{xia2023moderate} selects samples that have the closest distances to the class median feature $\boldsymbol{z}^{j}$,
% \begin{equation}
% \left\{\boldsymbol{z}^{j} = \frac{\sum_{i = 1}^{n} \mathbb{I}\left[y_{i} = j\right] \boldsymbol{z}_{i}}{\sum_{i = 1}^{n} \mathbb{I}\left[y_{i} = j\right]}\right\}_{j = 1}^{k},
% \label{eq:eq3}   
% \end{equation}%
where the class-wise mean of the embeddings is computed by averaging across all representation dimensions. 
To determine the distance from each representation to its corresponding class median, the scoring criterion is computed as
$S_{\mathrm{Moderate-DS}}(\boldsymbol{z}_i)=||\boldsymbol{z}_i - \boldsymbol{z}^{j}||^{2}$,   
which is the squared Euclidean distance  between embeddings $\{\boldsymbol{z}_1, \ldots,\boldsymbol{z}_n\}$ and class medians $\{\boldsymbol{z}^1, \ldots,\boldsymbol{z}^{k}\}$. 
Subsequently, with a given data ratio $R$, the coreset $D_s$ is defined as all data points within a distance of $\frac{R}{2}$ from the median.

\subsection{Class-wise Reweighting}
Distinct differences in sampling abilities persist across all classes, as shown in Fig.~\ref{fig:fig2}. Class-wise reweighting aims to improve overall generative performance after dataset pruning by considering these differences between diverse domains. To acquire class weights, a proxy model is trained by the worst-case loss~\cite{sagawa2020distributionally, xie2023doremi} over classes, which follows a mini-max optimization as distributionally robust optimization (DRO):
\begin{equation}
\displaystyle \min_{\boldsymbol \theta}\max_{ \boldsymbol{\alpha} \in \Delta }
\sum_{i=1}^K
\left [\boldsymbol{\alpha}\left ( 
\ell_i(\boldsymbol{\theta}; D_{s_i})
- \ell_\mathrm{ref}(\boldsymbol{\theta}_0; D_{s_i})
\right )
\right ],
%\cdot\sum_{j=1}^{|D_s|}\left[\ell_\theta(x_j)-\ell_{\text{ref}}(x_j)\right]
%
\label{eq:eq5}   
\end{equation}
where $K$ is the number of image classes, 
$\{\boldsymbol{ \alpha}_{1}, \ldots,\boldsymbol{ \alpha}_{K}\}$ signifies the corresponding class weights, $\Delta$ denotes the probability simplex constraint (\textit{i.e.}, $\boldsymbol{\alpha}_i > 0$ and $\sum_{i=1}^K \boldsymbol{\alpha}_i = 1$), 
$\ell_i(\boldsymbol{\theta}; D_{s_i})$ and $\ell_\mathrm{ref}(\boldsymbol{\theta}_0; D_{s_i}$) represent the loss of the proxy model and the reference model over the subset of images within class $i$  (denoted as $D_{s_i}$) respectively. 
The proxy and reference model share an identical architecture, whereas the proxy model is trainable on the selected subset and the reference model is pre-trained on the full training dataset. The margin loss $\ell_i(\cdot)-\ell_\mathrm{ref}(\cdot)$, which is only reserved to be greater than 0, quantifies the improvement space for the proxy model relative to the reference model on example $x$ from the coreset. Instances with higher excess loss are learnable and worth learning, where the reference model obtains a low loss, yet the proxy model still exhibits a high loss. 
% Conversely, examples with low excess loss may have either high entropy (resulting in a high learning loss and consequently a high reference loss) or low entropy (being easy to learn, leading to a low proxy loss). 
DRO adjusts class weights through gradient updates on the proxy model weights over training steps $t$, thereby amplifying the proxy model's gradient updating on some classes. The average class weight $\bar{\boldsymbol{\alpha}} = \frac{1}{T}\sum_{t=1}^T \boldsymbol{\alpha}^t$ over the $T$ training step is returned as the final class weight.

\section{Experiments}
\label{sec:exp}

\subsection{Experimental Setup} 

We vary the surrogate models as the encoder by Inceptionv3~\cite{szegedy2015rethinking}, ResNet-18~\cite{He_2016_CVPR}, ResNet-50-based CLIP~\cite{radford2021learning} and DDAE (an unconditional DDPM)~\cite{ddae2023} on three datasets with class labels, including CIFAR-10 with image size of 32$\times$32, ImageNet~\cite{5206848} and ImageNette (a subset containing 10 easy classes from ImageNet) with image size of 256$\times$256. The pixel-wise DM is DDPM~\cite{ho2020denoising} with classifier-free guidance~\cite{ho2022classifierfree} and LDMs are MDT~\cite{gao2023masked} with the size of S, mask ratio 0.3 and Adan optimizer~\cite{xie2024adan}, and SD~\cite{rombach2021highresolution}. DDPM is trained from scratch on 2000 epochs, MDT is trained on 60 epochs and SD is fine-tuned on 50 epochs. Generation quality is evaluated in Fr\'{e}chet Inception Distance (FID)~\cite{heusel2018gans} on 50k generated samples. Both DDPM and SD are efficiently inferenced by DDIM sampler~\cite{song2020denoising} with 100 and 50 steps, class classifier-free guidance~\cite{ho2022classifierfree} $w$ as 0.3 and 5 respectively.  MDT is evaluated with 250 DDPM sampling steps and 3.8 classifier-free guidance~\cite{ho2022classifierfree}. Considering a more reliable and unbiased estimator of image quality on ImageNet, we adopt CMMD~\cite{jayasumana2024rethinking} by Vision-Transformer-based CLIP embeddings and the maximum mean discrepancy distance with the Gaussian kernel. The DDPM proxy model on class reweighting follows the same setups mentioned above. The MDT proxy model follows similar settings except for 6 training epochs.

\begin{figure}[htbp]
  \centering
  % \fbox{\rule{0pt}{2in} \rule{1.0\linewidth}{0pt}}
   \includegraphics[width=1\linewidth]{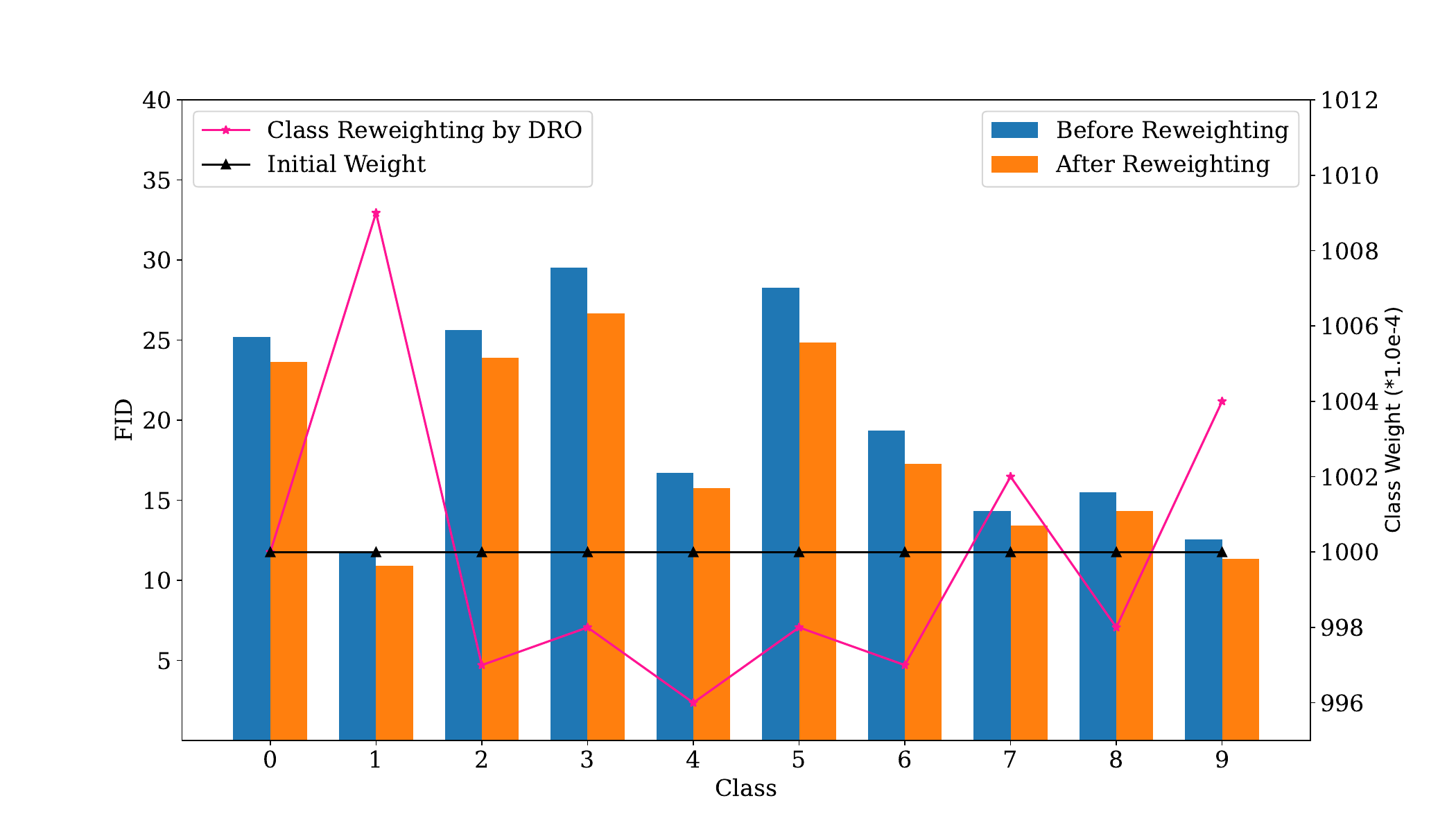}
   \caption{Class-wise DDPM FIDs (lower is better) before and after reweighting on 10\% of CIFAR-10 training data. Reweighting leads to improved generation performance across all 10 classes; the class weights larger than the initial weight are from the easily generated categories.}
   % however, differences in sampling abilities remain distinct even after reweighting.}
   \label{fig:fig2}
\end{figure}

\subsection{DDPM Results} 
We first investigate the data-efficient training of DMs on CIFAR-10 with low resolution (32$\times$32). In Table~\ref{tab:tab1}, DMs trained on pruned dataset size ranging from 5000 (10\% data ratio) to 20000 (40\%) show different generative capabilities. We discover that the GAN-based instance selection~\cite{devries2020instance} is generalized poorly to DMs, where FID scores are even greatly worse than those of Uniform Random, which selects data for each class randomly in a unified way. The performance decline is from 4.90 up to 13.91 with the decrease in dataset size, proving that it is difficult to describe the compact feature space by such an encoder and data pruning method. Therefore, we execute tentative experiments on surrogate encoding models, including ResNet-18~\cite{He_2016_CVPR}, CLIP~\cite{radford2021learning}, and DDAE (an unconditional self-supervised learner based on DDPM)~\cite{ddae2023} as image encoders. The scoring function is changed to Moderate-DS~\cite{xia2023moderate}, choosing those data samples close to the median. As shown in Table~\ref{tab:tab1}, ResNet-18 and DDAE yield more effective coresets compared to CLIP, which achieves lower FID than Uniform Random. The potential explanation, as supported by Table~\ref{tab:tab4}, is that both ResNet-18 and DDAE are trained solely on CIFAR-10 without any data transformations aimed at enlarging image size. In contrast, CLIP is pre-trained on a larger dataset with higher-resolution images. Consequently, latent features from CLIP on CIFAR-10 are somewhat expanded and sub-optimal. Another observation is that different classes maintain diverse generative abilities, as depicted in Fig.~\ref{fig:fig2}. To tackle this problem, class-wise reweighting is leveraged to enhance the generation levels from a class-specific perspective. Equipped with class-wise reweighting, DDPM trained on the coreset selected by ResNet-18 and Moderate-DS achieves a 6.71 FID score under 10\% of data and an FID score of 4.18 with 40\% of training samples. Inspired by InfoBatch~\cite{qin2024infobatch}, we consider a pruned dataset training ratio of 0.875 in Annealing to improve further generation abilities, where DDPM is trained on the subset before the 87.5\% training epoch and then on the full dataset until the end. We find that Annealing is significantly helpful in enhancing generative qualities especially when the dataset size is smaller and our approach outperforms InfoBatch~\cite{qin2024infobatch} in Table~\ref{tab:tab5}. Due to the partial full data training, the training speed is affected by Annealing in Table~\ref{tab:tab2}, showing the trade-offs between computational efficiency and image synthesis capacity.
\begin{table}[htbp]
  \caption{DDPM FID comparison results on 10\%-40\% of CIFAR-10 with annealing. Our method largely surpasses lossless training   acceleration framework InfoBatch~\cite{qin2024infobatch}.}
  \centering
  \resizebox{1.0\linewidth}{!}{
  \begin{tabular}{c|ccccc}
    \toprule[1pt]
          \multirow{2}*{Selection Method} & \multicolumn{4}{c}{FID ($\downarrow$) under Data Ratio}\\
     % & & \multicolumn{4}{c}{Data Ratio}\\ 
     & 10\% & 20\% & 30\%& 40\%\\
    \midrule
     InfoBatch + Annealing & 5.09  &4.28 &4.23 &4.04 \\
     Moderate-DS w/ Reweighting + Annealing &\textbf{4.58} &\textbf{4.19} &\textbf{4.12} &\textbf{3.92} \\
    \bottomrule[1pt]
  \end{tabular}
    }
  \label{tab:tab5}
\end{table}

\begin{table}[htbp]
  \caption{CIFAR-10 training speed-up on DDPM~\cite{ho2020denoising} by 10\%-40\% of data. The acceleration is up to \textbf{8.89} and reweighting has a minor impact on the training speed. However, annealing lowers the training acceleration because of few full-data training steps.}
  \centering
  \resizebox{1.0\linewidth}{!}{
  \begin{tabular}{c|c|cccc}
    \toprule[1pt]
         \multirow{2}*{Surrogate Model} & \multirow{2}*{Selection Method} & \multicolumn{4}{c}{Speed-Up under Data Ratio}\\
     % & & \multicolumn{4}{c}{Data Ratio}\\ 
     &  & 10\% & 20\% & 30\%& 40\%\\
    \midrule
    \multirow{3}*{ResNet-18} & Moderate-DS & 8.89 $\times$ &4.55 $\times$ &3.09 $\times$ &2.36 $\times$ \\
    & + Reweighting & 8.32 $\times$  &4.48 $\times$ &3.09 $\times$ &2.34 $\times$ \\
    & + Annealing & 4.38 $\times$  &3.14 $\times$ &2.45 $\times$ &2.02 $\times$ \\
    \bottomrule[1pt]
  \end{tabular}
    }
  \label{tab:tab2}
\end{table}

\subsection{MDT Results}
Our dataset pruning is further extended to MDT~\cite{gao2023masked} on ImageNet. MDT learns the contextual relation among object semantic parts by masking certain tokens in the latent space. Gaussian model selects a more superior and compact subset than Moderate-DS via feature embedding from Inceptionv3. Furthermore, class-wise reweighting is general to all data pruning approaches and different pruning ratios. Remarkably, as shown in Table~\ref{tab:tab3}, MDT equipped with reweighting on merely 20\% of data samples achieves a better FID score (13.94 vs. 17.11) than the model trained on the entire dataset. It demonstrates the possibility of both saving training costs and achieving qualified class-conditional image generation.

\begin{table}[htbp]
  \caption{ImageNet 256$\times$256 FID and CMMD evaluations for class-conditional MDT~\cite{gao2023masked} on [10\%, 20\%, 100\%] of training set. MDT trained on a subset could outperform the model trained on the full dataset.} 
  \centering
   \resizebox{1\linewidth}{!}{
  \begin{tabular}{c|c|cccc}
    \toprule[1pt]
     \multirow{2}*{Surrogate Model} & \multirow{2}*{Selection Method} & \multicolumn{3}{c}{FID / CMMD ($\downarrow$) under Data Ratio}\\ 
     & & 10\%  & 20\% & 100\%\\
    %  \toprule
    % Surrogate Model & Selection Method & \multicolumn{2}{c}{FID ($\downarrow$)} \\
    \midrule
    N/A &Uniform Random  &130.81 / 2.55
 & 51.18 / 1.45 &\multirow{5}*{17.11 / 0.86
}\\
    \multirow{4}*{Inceptionv3} & Gaussian & 95.68 / 2.44 &18.49 / 1.02  & \\
    & + Reweighting & \textbf{85.74} / \textbf{2.22} &\textbf{13.94} / \textbf{0.91} &\\
    &Moderate-DS &123.60 / 2.51 &50.63 / 1.46\\
   & + Reweighting & 115.62 / 2.46 &47.85 / 1.43\\
    \bottomrule[1pt]
  \end{tabular}
  }
  \label{tab:tab3}
\end{table}

\subsection{SD Results}
We evaluate dataset pruning on ImageNette, a subset from ImageNet, by fine tuning SD (Stable Diffusion ‘v1-4’~\cite{rombach2021highresolution}). The prompt for sampling is ‘a photo of a \textit{class name}'. By computing FID on a total of 50k sampling images in 256$\times$256 resolution under classifier-free guidance, SD fine-tuned on only 40\% of the data significantly surpasses the model on the entire dataset in Table~\ref{tab:tab4}. An interesting finding is that all models fine-tuned on the subsets show even better generation capability, highlighting the potential data redundancy in large LDM fine-tuning. Note that class-wise reweighting is not applied in this case, because SD fine-tuned on the subset has surpassed the one on the entire dataset, and thus majority of margin loss from Eq.~\eqref{eq:eq5} is clipped to 0.

\begin{table}[htbp]
  \caption{ImageNette 256$\times$256 FID evaluations for classifier-free guided SD~\cite{rombach2021highresolution} on [20\%, 40\%, 60\%, 100\%] of training set. Images of higher quality are generated by a DM fine-tuned on fewer data samples.  }
  \centering
  \resizebox{1\linewidth}{!}{
  \begin{tabular}{c|c|cccc}
    \toprule[1pt]
     \multirow{2}*{Surrogate Model} & \multirow{2}*{Selection Method} & \multicolumn{4}{c}{FID ($\downarrow$) under Data Ratio}\\ 
     & & 20\%  & 40\% & 60\% & 100\%\\
    %  \toprule
    % Surrogate Model & Selection Method & \multicolumn{2}{c}{FID ($\downarrow$)} \\
    \midrule
    N/A &Uniform Random  & 20.33 &21.21 & 23.44 &\multirow{3}*{25.91}\\
    CLIP  &\multirow{2}*{Moderate-DS} &20.27 &20.20 &23.19\\
    ResNet-18  &  &\textbf{20.05} &\textbf{18.51} &\textbf{20.94}\\
    \bottomrule[1pt]
  \end{tabular}
  }
  \label{tab:tab4}
\end{table}

\section{Conclusion}
\label{sec:concl}
In this work, we investigate data-efficient DM training by data selection and class reweighting.
As the first study on data-pruned DM training, we demonstrate its remarkable robustness across various, reducing computing overhead by up to 8$\times$. Furthermore, we reveal the presence of training data redundancy in both pixel-level and latent-level DMs. Overall, we believe our findings and approach provide a solid foundation for building
scalable and efficient artificial intelligence-generated content systems. 

\newpage
{{
\bibliographystyle{IEEEtran}
\bibliography{refs}
}}
\end{document}